\documentclass[11pt]{article}
\usepackage[a4paper,margin=27mm]{geometry}
\usepackage[T1]{fontenc}
\usepackage{lmodern}
\usepackage{amsmath,amssymb,amsthm,mathtools}
\usepackage{microtype}
\usepackage{needspace}
\usepackage[authoryear,round]{natbib}
\usepackage[hidelinks]{hyperref}
\usepackage{aliascnt}
\usepackage{cleveref}
\allowdisplaybreaks[1]
\hypersetup{pdftitle={Lower bounds on inference complexity for averaged stochastic gradient descent},pdfauthor={Denis Belomestny, Alexey Naumov, Sergey Samsonov}}
\newcommand{\PE}{\mathbb{E}}
\newcommand{\PP}{\mathbb{P}}
\newcommand{\rset}{\mathbb{R}}

\newcommand{\var}{\operatorname{Var}}
\newcommand{\Cov}{\operatorname{Cov}}
\newcommand{\filtr}{\mathcal{F}}
\newcommand{\prtheta}{\bar{\theta}}
\newcommand{\norm}[1]{\left\Vert#1\right\Vert}
\newcommand{\dk}{d_{\mathrm K}}
\newcommand{\dc}{d_{\mathcal C}}
\newcommand{\ph}{\varphi}
\newtheorem{theorem}{Theorem}[section]
\crefname{theorem}{theorem}{theorems}
\Crefname{theorem}{Theorem}{Theorems}
\newaliascnt{lemma}{theorem}
\newtheorem{lemma}[lemma]{Lemma}
\aliascntresetthe{lemma}
\crefname{lemma}{lemma}{lemmas}
\Crefname{lemma}{Lemma}{Lemmas}

\newtheorem{assumption}{\textbf{A}\hspace{-3pt}}
\Crefname{assumption}{\textbf{A}\hspace{-3pt}}{\textbf{A}\hspace{-3pt}}
\crefname{assumption}{\textbf{A}}{\textbf{A}}

\theoremstyle{remark}

\title{Lower Bounds on Inference Complexity for Polyak-Ruppert Averaged Stochastic Gradient Descent}
\author{Denis Belomestny~\footnote{University of Duisburg-Essen, Germany,
and HSE University, Moscow, Russia.}
\and Alexey Naumov~\footnote{HSE University, Moscow, Russia.}
\and Sergey Samsonov~\footnote{HSE University, Moscow, Russia,  \texttt{svsamsonov@hse.ru}.}}
\date{}
\begin{document}
\maketitle
\begin{abstract}
In this paper, we establish sharp lower bounds for Gaussian approximation of
Polyak-Ruppert averaged stochastic gradient descent with polynomially
decreasing step sizes. For every fixed exponent $\gamma\in(1/2,1)$,
a single one-dimensional smooth, strongly convex example has
approximation error with the limiting variance of order
$n^{1/2-\gamma}+n^{\gamma-1}$ in Kolmogorov and convex distances. This matches the existing upper
bound throughout the range of $\gamma$. The nonlinear bias contributes
$n^{1/2-\gamma}$ even when the Gaussian reference uses the finite
linearized variance. We also derive a lower bound of order
$n^{-(1-\gamma)/2}$ for the relative mean error of estimating the
limiting variance from one SGD trajectory, uniformly over a scalar
quadratic family. With the limiting variance known, achieving approximation error
$\varepsilon$ on this example takes order $\varepsilon^{-4}$
iterations for the best choice $\gamma=3/4$. Achieving both Gaussian approximation and
relative mean variance-estimation errors at most $\varepsilon$
uniformly over the model class requires at least order
$\varepsilon^{-6}$ iterations for every fixed $\gamma$.
\end{abstract}

\section{Introduction}
\label{sec:intro}
We study lower bounds for Gaussian approximation of Polyak--Ruppert
averaged stochastic gradient descent (SGD). Let
$f:\rset^d\to\rset$ be a smooth, strongly convex function with
minimizer $\theta^\star$. Given independent and identically
distributed observations $(Z_k)_{k\ge1}$ and an unbiased stochastic
gradient $\PE F(\theta,Z_1)=\nabla f(\theta)$, SGD generates
\begin{equation}
\label{eq:iterates_SGD}
 \theta_k=\theta_{k-1}-\alpha_k F(\theta_{k-1},Z_k).
\end{equation}
The Polyak--Ruppert average is
\begin{equation}
\label{eq:pr_theta_def}
 \prtheta_n=\frac1n\sum_{k=0}^{n-1}\theta_k.
\end{equation}
Under suitable regularity, moment, and step-size conditions,
\citet{ruppert1988} and \citet{polyak1992} establish
\begin{equation}\label{eq:intro-clt}
 \sqrt n(\prtheta_n-\theta^\star)
 \xrightarrow{\mathrm d}N(0,\Sigma_\infty),
 \qquad \Sigma_\infty=G^{-1}VG^{-1},
\end{equation}
where $G=\nabla^2 f(\theta^\star)$ and
$V=\Cov(F(\theta^\star,Z_1))$.
To quantify this convergence in dimension $d$, let
$Y_\infty\sim N(0,\Sigma_\infty)$ and write
\[
 \Delta_n=\sup_{B\in\mathcal C(\rset^d)}
 \left|\PP\bigl(\sqrt n(\prtheta_n-\theta^\star)\in B\bigr)
          -\PP(Y_\infty\in B)\right|,
\]
where $\mathcal C(\rset^d)$ is the collection of convex subsets of
$\rset^d$. Consider polynomial step sizes
$\alpha_k=c_0(k+k_0)^{-\gamma}$, with $1/2<\gamma<1$ and suitable
constants $c_0>0$ and $k_0\ge1$. Under smoothness and moment
assumptions, \citet[Theorem~4]{sgd} prove the Gaussian approximation
bound
\begin{equation}\label{eq:intro-upper}
 \Delta_n\le C\left\{
 n^{-1/2}+n^{1/2-\gamma}+n^{-\gamma/2}+n^{\gamma-1}
 \right\},
\end{equation}
where $C$ is independent of $n$.
The two leading terms are $n^{1/2-\gamma}$ and $n^{\gamma-1}$,
and balancing them gives the rate $n^{-1/4}$ at $\gamma=3/4$.
Their quadratic example proves that the term $n^{\gamma-1}$ is
necessary, establishing sharpness for $\gamma\ge3/4$.
The necessity of the term $n^{1/2-\gamma}$ for
$1/2<\gamma<3/4$ remains open in that work.

For a fixed $a\in\rset^d$ with
$\sigma_a^2=a^\top\Sigma_\infty a>0$, the asymptotic normality \eqref{eq:intro-clt} suggests the following confidence interval for
the linear contrast $a^\top\theta^\star$ at nominal level
$1-\delta$, where $0<\delta<1$:
\begin{equation}\label{eq:intro-wald}
 C_{n,\delta}^{\mathrm{or}}(a)
 =\left[
 a^\top\prtheta_n-z_{1-\delta/2}\frac{\sigma_a}{\sqrt n},\;
 a^\top\prtheta_n+z_{1-\delta/2}\frac{\sigma_a}{\sqrt n}
 \right],
\end{equation}
where $z_p=\Phi^{-1}(p)$ and $\Phi$ is the standard normal
distribution function. The superscript indicates that the limiting
covariance is treated as known. The coverage event corresponds to a
convex slab for $\sqrt n(\prtheta_n-\theta^\star)$, so
\begin{equation}\label{eq:intro-coverage-bound}
 \left|\PP\bigl(a^\top\theta^\star\in
 C_{n,\delta}^{\mathrm{or}}(a)\bigr)-(1-\delta)\right|
 \le\Delta_n.
\end{equation}
Thus \eqref{eq:intro-upper} gives a finite-sample guarantee for the
coverage of \eqref{eq:intro-wald}. At $\gamma=3/4$, order
$\varepsilon^{-4}$ iterations suffice to make this coverage error
at most $\varepsilon$, with the problem parameters fixed.
The interval's width decreases as $n^{-1/2}$. The guaranteed coverage
error decreases as $n^{-1/4}$ for this schedule.

Lower bounds give necessary iteration counts for Gaussian
approximation. A sharp approximation bound therefore
determines the iteration complexity of this uniform calibration
requirement. In practice, $\Sigma_\infty$ is unknown, and a plug-in interval
replaces $\sigma_a$ in \eqref{eq:intro-wald} by
$(a^\top\widehat\Sigma_n a)^{1/2}$.
Its accuracy also depends on the estimation of this scale.
Online covariance estimators allow this quantity to be updated
during SGD \citep{chen2020,ZhuChenWu2023,WeiZhuLyuWu2026}.
\citet{NiHuo2026} obtain a covariance-estimation lower bound by
comparing trajectories generated by quadratic objectives with nearby
curvatures. We combine this approach with a sharp Gaussian
approximation lower bound to study how many iterations are necessary
to control both errors from a single SGD trajectory.

Our main contributions are as follows.
\begin{enumerate}
 \item We construct a one-dimensional smooth, strongly convex
 objective and an unbiased stochastic gradient oracle for which
 the Gaussian approximation error with the limiting variance is of order
 $n^{1/2-\gamma}+n^{\gamma-1}$ in both Kolmogorov and convex
 distances (\Cref{thm:nonlinear}). The same example works for
 every fixed $\gamma\in(1/2,1)$ and proves the sharpness of
 \eqref{eq:intro-upper}. The term $n^{1/2-\gamma}$ comes from
 nonlinear bias and persists when the Gaussian reference uses the
 finite linearized variance.

 \item We adapt the construction of \citet{NiHuo2026} to a scalar
 quadratic family and the step sizes used in our example.
 Over this family, the worst-case relative mean error of every
 estimator of the limiting variance based on one trajectory is at
 least of order $n^{-(1-\gamma)/2}$
 (\Cref{thm:covariance-lower}). The estimator observes the iterates
 and step sizes, with no access to raw observations, stochastic
 Hessians, or additional gradient evaluations.

 \item We combine the two lower bounds to show that simultaneous
 Gaussian approximation error and relative mean variance-estimation
 error at most $\varepsilon$ require
 \[
 n\gtrsim
 \varepsilon^{-\max\{1/(\gamma-1/2),\,2/(1-\gamma)\}}
 \]
 uniformly over the model class
 (\Cref{thm:inference-complexity}). The smallest exponent in this
 lower bound is six, at $\gamma=2/3$.
 For Gaussian approximation with known limiting variance, the
 smallest complexity exponent on our example is four, at
 $\gamma=3/4$.
\end{enumerate}
These complexity statements concern the fixed polynomial schedules
studied here. The joint requirement controls Gaussian approximation
and variance estimation separately; coverage after studentization
also depends on their joint distribution. We give a sufficient
coverage bound allowing both quantities to be computed from the
same trajectory in \Cref{lem:studentization-transfer}.

\subsection{Related work}\label{sec:related}
\paragraph{Averaging and optimization accuracy.}
Stochastic approximation and its asymptotic normality originate in
\citet{robbins1951} and \citet{fabian1968}.
Polyak--Ruppert averaging achieves the statistical $n^{-1/2}$ estimation
scale under suitable conditions \citep{ruppert1988,polyak1992}.
Non-asymptotic analyses quantify its optimization error
\citep{nemirovski2009,moulines2011,BachMoulines2013,Bach2014}.
\citet{GadatPanloup2023} identify the limiting covariance in the leading
mean-square term and obtain a second-order upper bound optimized at
$\gamma=3/4$. Our Gaussian approximation bound selects the same
exponent for a distributional accuracy criterion.
Weighted averaging and general learning-rate schedules are studied by
\citet{WeiZhuWu2026} and \citet{GoldreichEtAl2025}.

\paragraph{Gaussian approximation for stochastic approximation.}
\citet{shao2022} prove Berry-Esseen bounds for nonlinear statistics
and apply them to averaged SGD.
\citet{sgd} compare the normalized average with Gaussian laws having
either the finite linearized covariance or its limit. Their quadratic lower bound identifies the
error caused by replacing the former covariance by the latter.
Our nonlinear example also identifies a bias contribution that
persists with the finite linearized variance.
The proof uses the perturbation expansion developed for linear
stochastic approximation by \citet{durmus}, together with a curvature
correction and a comparison of smooth test expectations.

For linear stochastic approximation with independent noise,
\citet{SamsonovEtAl2024} establish multivariate Gaussian approximation
and multiplier-bootstrap bounds.
\citet{ButyrinEtAl2025} improve the Gaussian approximation rate to
$n^{-1/3}$, up to logarithmic factors, for a suitable step-size
schedule. Gaussian approximation for linear functionals of online
least-squares SGD in growing dimension is studied by
\citet{agrawalla2026}.
For Markovian noise, \citet{SamsonovEtAl2025Markov} obtain Gaussian
approximation and block-bootstrap guarantees.
More recently, \citet{CaiQinZhang2026} study scalar linear stochastic
approximation with Markovian noise. Using the finite linearized variance for normalization, they obtain
a Berry--Esseen rate $n^{-\gamma/2}$ after deterministic bias
correction and an asymptotic error of order $n^{1/2-\gamma}$ for the
uncorrected statistic when the leading bias coefficient is nonzero. Their bias arises in a linear recursion with dependent noise;
our construction uses nonlinear curvature and independent noise.
\citet{LiLiangZhang2026} study Brownian approximation of fixed scalar
projections of the normalized partial-sum process in Prokhorov
distance. They give separate lower-bound constructions for covariance
stabilization, finite-moment noise, algorithmic smoothing, and
nonlinearity. Our bounds concern the Kolmogorov and convex distances
for the terminal averaged estimate on one fixed smooth example.

\paragraph{Constant step sizes.}
The stationary distribution of constant-step SGD has been studied in
\citet{Pflug1986}, \citet{dieuleveut2020}, and
\citet{ChenMouMaguluri2022}.
Recent quantitative results include convergence and concentration
bounds \citep{MeradGaiffas2025}, Gaussian approximation under constant
learning rates \citep{WeiLiLouWu2025,WangEtAl2026}, and convex-distance
bounds for online linear regression
\citep{SheshukovaDurmusKhusainovSamsonov2026}.
\citet{Seo2026} proves an $O(\sqrt{\alpha})$ approximation in
$1$-Wasserstein distance for the normalized stationary iterate of
SGD with constant step size $\alpha$ and bounded additive Markovian
noise. A four-state example gives a matching lower bound.
\citet{LiWu2026Stationary} study stationary Gaussian limits for least-squares
SGD with dependent designs. Under stronger design assumptions, they
identify a positive first-order term in total variation and a
deterministic covariance correction. These stationary results
complement the lower bounds for decreasing-step averages studied here.

\paragraph{Covariance estimation and confidence intervals.}
\citet{chen2020} develop plug-in and batch-means covariance estimators
for SGD. Online estimators and their refinements are studied by
\citet{ZhuChenWu2023}, \citet{SinghShuklaVats2025}, and
\citet{WeiZhuLyuWu2026}.
\citet{NiHuo2026} prove a trajectory-based covariance-estimation lower
bound using quadratic objectives with nearby curvatures. We adapt
their construction to one dimension and our fixed step-size schedule,
then combine it with the Gaussian approximation lower bound.
The estimator uses only the iterates and known step sizes.

Other inference procedures use random scaling
\citep{LeeLiaoSeoShin2022}, hierarchical or parallel SGD runs
\citep{SuZhu2023,ZhuLouWeiWu2024}, and the bootstrap
\citep{fang2018,sgd,LamWang2026}.
Inference for implicit SGD is studied by
\citet{Fang2019} and \citet{LeeLeeWon2022}.
Concentration inequalities provide another way to construct
confidence bounds \citep{ZhuLouWu2022}.
\paragraph{Information lower bounds.}
The oracle framework of \citet{NemirovskyYudin1983} quantifies
optimization complexity through the information available in oracle
responses. Stochastic first-order lower bounds are developed by
\citet{AgarwalEtAl2012} and \citet{RaginskyRakhlin2011}, while
\citet{ZhuEtAl2016} and \citet{DuchiRuan2021} study local minimax complexity and
asymptotic efficiency. Our covariance lower bound also uses a
two-point testing argument. The observations are generated by the
fixed SGD recursion, and the unknown curvature determines the
limiting variance. The lower bound ranges over estimators of this
variance from the observed trajectory. Our Gaussian approximation
lower bound concerns the distribution of the average on a fixed
objective.

\paragraph{Notation.}
For real random variables $X$ and $Y$, define
\begin{align*}
 \dk(X,Y)&=\sup_{t\in\rset}
       |\PP(X\le t)-\PP(Y\le t)|,\\
 \dc(X,Y)&=\sup_{I\subset\rset\,\mathrm{convex}}
       |\PP(X\in I)-\PP(Y\in I)|.
\end{align*}
Then $\dk\le\dc\le2\dk$. Let $\Phi$ and $\ph$ be the standard normal
 distribution function and density, and set
$\norm{V}_p=(\PE|V|^p)^{1/p}$.
The notation $a_n\asymp b_n$ means that their ratio is eventually
bounded above and below by positive constants.
We write $a_n\gtrsim b_n$ when $a_n\ge c b_n$ eventually for some
constant $c>0$ independent of $n$.

\section{Main result}

We work on $\rset$ and translate the minimizer to zero. We first
impose strong convexity and regularity conditions on the objective.

\begin{assumption}\label{ass:objective}
The function $f:\rset\to\rset$ is twice continuously differentiable
and $f'(0)=0$. There exist positive constants $\mu,L_1,L_3$ such that,
for all $x,y\in\rset$,
\begin{align}
 \mu&\le f''(x)\le L_1,\label{eq:ass-convexity}\\
 |f''(x)-f''(y)|&\le L_3|x-y|.\label{eq:ass-hessian}
\end{align}
\end{assumption}

The stochastic gradient has a centered additive noise component and
a state-dependent component satisfying a moment Lipschitz condition.

\begin{assumption}\label{ass:oracle}
The observations $(Z_k)_{k\ge1}$ are independent and identically
distributed. The oracle has the representation
\begin{equation}\label{eq:ass-oracle}
 F(x,Z)=f'(x)+\eta(Z)+g(x,Z),
\end{equation}
where
\begin{align*}
 \PE\eta(Z)&=0,\\
 0<\PE\eta(Z)^2&<\infty,\\
 \PE|\eta(Z)|^4&<\infty,\\
 \PE g(x,Z)&=0,\qquad g(0,Z)=0.
\end{align*}
There exists $L_2<\infty$ such that, for every $x,y\in\rset$ and
$2\le p\le4$,
\begin{equation}\label{eq:ass-noise-lipschitz}
 \norm{g(x,Z)-g(y,Z)}_p\le L_2|x-y|.
\end{equation}
\end{assumption}

We use polynomially decreasing step sizes and initialize the recursion
at the minimizer, as specified next.

\begin{assumption}\label{ass:polynomial}
For fixed $1/2<\gamma<1$, $k_0\ge1$, and $c_0>0$ satisfying
$2c_0L_1\le1$, let
\begin{equation}\label{eq:ass-polynomial}
 \alpha_k=c_0(k+k_0)^{-\gamma}.
\end{equation}
Initialize at $\theta_0=0$ and define
\begin{equation}\label{eq:sgd-update}
 \theta_k=\theta_{k-1}-\alpha_k F(\theta_{k-1},Z_k).
\end{equation}
\end{assumption}

Note that 
\Cref{ass:objective} gives strong convexity, a Lipschitz gradient, and
a globally Lipschitz Hessian. The last condition is stronger than the
local Hessian condition in \citet{sgd}. The fixed example below satisfies
A1, A3, A7(4), and A8 of their Theorem~4. In particular, its
noise Lipschitz bound holds for every $p\ge2$, including the
range $2\le p\le\log n$ in their A2(ii). The noise in
\Cref{ass:oracle} may grow with the distance from the minimizer;
a uniform bound on $F(x,Z)-f'(x)$ over $x$ is not required.

\subsection{The example}
Let $(R_k,S_k)_{k\ge1}$ be independent pairs, with $R_k$ and $S_k$
independent Rademacher random variables. We use the fixed objective
and oracle
\begin{align}
\label{eq:nonlinear-example}
f(x) = \frac{x^2}{2}+\frac{x-\sin x}{2}, \quad F(x;R,S) = (1+R)f'(x) + S\,.
\end{align}
The recursion is
\begin{align}
\label{eq:nonlinear-recursion}
\theta_k = \theta_{k-1}-\alpha_kF(\theta_{k-1};R_k,S_k),
 \quad \theta_0=0, \quad \text{ where } \alpha_k =\frac{1}{4(k+1)^\gamma},
 \quad \frac12<\gamma<1\,.
\end{align}

We verify the assumptions directly. Differentiation gives
\begin{align}
\label{eq:gradient-example}
 f'(x) = x+\frac{1-\cos x}{2}, \quad 
 f''(x) =1+\frac12\sin x\,.
\end{align}
Hence $f'(0)=0$ and
\begin{align*}
 \frac12 \le f''(x)\le\frac32, \quad |f''(x)-f''(y)|&\le\frac12|x-y|.
\end{align*}
The unique minimizer is therefore zero. Since $\PE R=\PE S=0$,
the oracle is unbiased, and its noise decomposition is
\[
 \eta(R,S)=S,\qquad g(x;R,S)=Rf'(x).
\]
In particular,
\begin{align*}
 \PE\eta^2&=\PE\eta^4=1,\\
 \norm{g(x;R,S)-g(y;R,S)}_p
 &=|f'(x)-f'(y)|
 \le\frac32|x-y|\qquad(p\ge2).
\end{align*}
Thus \Cref{ass:objective,ass:oracle} hold with
$\mu=1/2$, $L_1=L_2=3/2$, and $L_3=1/2$.
For \eqref{eq:nonlinear-recursion}, $c_0=1/4$, $k_0=1$, and
$2c_0L_1=3/4<1$, so \Cref{ass:polynomial} also holds.
The limiting Gaussian variance of the normalized average is one.

\subsection{Lower bounds for Polyak--Ruppert averaging}\label{sec:pr}
In this section the step sizes satisfy \Cref{ass:polynomial}, with the
fixed choice in \eqref{eq:nonlinear-recursion}. First, denote
\begin{align}
P_{i,k}=\prod_{\ell=i+1}^k(1-\alpha_\ell), \quad 0\le i\le k, \qquad Q_{i,n}=\alpha_i\sum_{k=i}^{n-1}P_{i,k}, \quad 1\le i<n.
\label{eq:weights}
\end{align}
Empty products equal one. We consider
\begin{align}
 X_n&=\sqrt n\prtheta_n
      =\frac1{\sqrt n}\sum_{k=0}^{n-1}\theta_k,
      &Z&\sim N(0,1),\label{eq:pr-statistic}\\
 s_n^2&=\frac1n\sum_{i=1}^{n-1}Q_{i,n}^2,
      &Z_n&\sim N(0,s_n^2).\label{eq:pr-linear-variance}
\end{align}
Here $s_n^2$ is the variance of the linearized average. It is not the
variance of $X_n$.

The next theorem gives sharp Gaussian approximation lower bounds
for the averaged iterate, comparing it with both the limiting Gaussian
and the Gaussian with the finite linearized variance.

\begin{theorem}
\label{thm:nonlinear}
For the example \eqref{eq:nonlinear-example} with recursion
\eqref{eq:nonlinear-recursion},
\begin{equation}
 \liminf_{n\to\infty}
 \frac{\dk(X_n,Z)}
 {\displaystyle\ph(1)\left\{
       \frac{n^{1/2-\gamma}}{32(1-\gamma)}+n^{\gamma-1}\right\}}
 \ge1.
 \label{eq:nonlinear-lower}
\end{equation}
where $\varphi$ is the standard Gaussian density. 
The same lower bound holds with $\dc$ in place of $\dk$. In particular,
\begin{equation}
 \dk(X_n,Z)\asymp\dc(X_n,Z)
 \asymp n^{1/2-\gamma}+n^{\gamma-1}.
 \label{eq:nonlinear-order}
\end{equation}
For the Gaussian law with the finite linearized variance,
\begin{align}
 \liminf_{n\to\infty}n^{\gamma-1/2}\dk(X_n,Z_n)
 &\ge\frac{\ph(0)}{32(1-\gamma)},\label{eq:pr-finite-lower}\\
 \dk(X_n,Z_n)\asymp\dc(X_n,Z_n)
 &\asymp n^{1/2-\gamma}.\label{eq:pr-finite-order}
\end{align}
The comparison constants may depend on $\gamma$.
\end{theorem}

The upper bound in \eqref{eq:nonlinear-order} follows from
\citet[Theorem~4]{sgd}. Its additional terms $n^{-1/2}$ and $n^{-\gamma/2}$ are smaller
than $n^{1/2-\gamma}$. Similarly, \citet[Theorem~2]{sgd} gives the
upper bound in \eqref{eq:pr-finite-order}: the additional terms
$n^{-1/2}$ and $n^{-\gamma/2}$ are smaller than $n^{1/2-\gamma}$.
We prove the lower bounds below. They arise from two effects with
nonzero coefficients: the mean of the curvature correction and the
difference between $s_n^2$ and its limit.

\begin{proof}[Proof of \Cref{thm:nonlinear}]
Fix $\tau>0$ and set
\[
 a_\tau=\sqrt{1+\tau^2},\qquad
 h_\tau(x)=\Phi\left(\frac{-a_\tau-x}{\tau}\right).
\]
This function is decreasing, takes values in $[0,1]$, and has total
variation one. Integration against the distribution functions gives
\begin{equation}
 |\PE h_\tau(V)-\PE h_\tau(V')|
 \le\dk(V,V')
 \label{eq:pr-smooth-test}
\end{equation}
for any real random variables $V,V'$. Its Lipschitz constant is at
most $1/(\tau\sqrt{2\pi})$. We first compare $X_n$ with $W_n-B_n$, where
\[
 W_n=-\frac1{\sqrt n}\sum_{i=1}^{n-1}Q_{i,n}S_i.
\]
and $B_n$ is defined in \eqref{eq:pr-bias-definition}. By \Cref{lem:pr-reduction},
\begin{align*}
 |\PE h_\tau(X_n)-\PE h_\tau(W_n-B_n)|
 =o(n^{1/2-\gamma}).
\end{align*}
Next, \eqref{eq:pr-smooth-test} and \eqref{eq:pr-weighted-be} imply
\begin{align*}
 |\PE h_\tau(W_n-B_n)-\PE h_\tau(Z_n-B_n)|
 &\le\dk(W_n,Z_n)\\
 &\le Cn^{-1/2}
 =o(n^{1/2-\gamma}).
\end{align*}
The last relation uses $\gamma<1$. Gaussian convolution now gives
\begin{align}
 \PE h_\tau(X_n)
 &=\Phi\left(
       \frac{-a_\tau+B_n}{\sqrt{\tau^2+s_n^2}}
       \right)+o(n^{1/2-\gamma}),\label{eq:pr-smoothed-cdf}\\
 \PE h_\tau(Z)&=\Phi(-1).\nonumber
\end{align}
Write $\delta_n=s_n^2-1$. By \Cref{lem:covariance},
$\delta_n=2n^{\gamma-1}+o(n^{\gamma-1})$. Lemma \ref{lem:pr-bias} implies that $B_n\to0$. Since
$\delta_n\to0$, a Taylor expansion of the argument in
\eqref{eq:pr-smoothed-cdf} yields
\begin{align*}
 \frac{-a_\tau+B_n}{\sqrt{\tau^2+s_n^2}}
 &=\left(-1+\frac{B_n}{a_\tau}\right)
      \left(1+\frac{\delta_n}{a_\tau^2}\right)^{-1/2}\\
 &=-1+\frac{B_n}{a_\tau}
          +\frac{\delta_n}{2a_\tau^2}
          +o(B_n+|\delta_n|).
\end{align*}
Applying a second Taylor expansion, now to $\Phi$, and using
\Cref{lem:pr-bias}, we obtain
\begin{align}
 \PE h_\tau(X_n)-\PE h_\tau(Z)
 &=\ph(1)\left(
       \frac{B_n}{a_\tau}
       +\frac{\delta_n}{2a_\tau^2}\right)
       +o(n^{1/2-\gamma}+n^{\gamma-1})\nonumber\\
 &=\ph(1)\left\{
       \frac{n^{1/2-\gamma}}{32(1-\gamma)a_\tau}
       +\frac{n^{\gamma-1}}{a_\tau^2}\right\}
       +o(n^{1/2-\gamma}+n^{\gamma-1}).
 \label{eq:pr-two-effects}
\end{align}
Both coefficients in braces are positive. For this fixed $\tau$,
\eqref{eq:pr-smooth-test} and \eqref{eq:pr-two-effects} imply that
the lower limit in \eqref{eq:nonlinear-lower} is at least
$\min(a_\tau^{-1},a_\tau^{-2})$. Letting $\tau\downarrow0$ proves
\eqref{eq:nonlinear-lower}. Since $\dc\ge\dk$, the same lower
bound holds for convex distance. The previously cited upper bound
then proves \eqref{eq:nonlinear-order}.

For the finite linearized covariance, use instead
\[
 \widetilde h_\tau(x)=\Phi(-x/\tau).
\]
The expectation of this function under any centered Gaussian law
is $1/2$. Repeating the preceding two comparison steps gives
\begin{align*}
 \PE\widetilde h_\tau(X_n)-\PE\widetilde h_\tau(Z_n)
 &=\Phi\left(\frac{B_n}{\sqrt{\tau^2+s_n^2}}\right)
      -\frac12+o(n^{1/2-\gamma})\\
 &=\frac{\ph(0)B_n}{\sqrt{\tau^2+s_n^2}}
      +o(n^{1/2-\gamma}).
\end{align*}
It follows that
\[
 \liminf_{n\to\infty}n^{\gamma-1/2}\dk(X_n,Z_n)
 \ge\frac{\ph(0)}{32(1-\gamma)\sqrt{1+\tau^2}}.
\]
Letting $\tau\downarrow0$ proves \eqref{eq:pr-finite-lower}.
Together with the cited finite-variance upper bound, this proves
\eqref{eq:pr-finite-order}.
\end{proof}

\section{Lower bounds on inference complexity}\label{sec:calibration}
The Gaussian approximation in \Cref{thm:nonlinear} uses a deterministic
reference variance. When this variance is unknown, a plug-in Gaussian procedure also
requires a variance estimate. We now consider these two accuracy requirements together.
The covariance lower bound depends on the information available to the
estimator, so we specify the observation model before stating the result.

For a scalar model $M$ satisfying \Cref{ass:objective,ass:oracle}, write
\begin{equation}\label{eq:limiting-variance}
 \Sigma_\infty(M)
 =\frac{\PE_M\eta(Z)^2}{f_M''(\theta_M^\star)^2}.
\end{equation}
This is the limiting variance of the normalized Polyak--Ruppert
average. For the nonlinear example in \eqref{eq:nonlinear-example},
$\Sigma_\infty=1$. It differs from the finite linearized variance
$s_n^2$ in \eqref{eq:pr-linear-variance}.

\subsection{Gaussian calibration with known variance}
Let
\begin{equation}\label{eq:calibration-error}
 \Delta_{n,M}
 =\dk\bigl(\sqrt n(\prtheta_n-\theta_M^\star),
                   N(0,\Sigma_\infty(M))\bigr).
\end{equation}
For $p\in(0,1)$, consider the one-sided Gaussian confidence set
\begin{equation}\label{eq:confidence-set}
 I_{n,p}
 =\left[\prtheta_n-
       \sqrt{\frac{\Sigma_\infty(M)}n}\,\Phi^{-1}(p),\infty\right).
\end{equation}
Its coverage event is
\begin{align*}
 \{\theta_M^\star\in I_{n,p}\}
 &=\left\{\sqrt n(\prtheta_n-\theta_M^\star)
       \le\sqrt{\Sigma_\infty(M)}\,\Phi^{-1}(p)\right\}.
\end{align*}
As $p$ ranges over $(0,1)$, the threshold ranges over $\rset$.
Consequently,
\begin{equation}\label{eq:coverage}
 \sup_{0<p<1}
 \left|\PP_M(\theta_M^\star\in I_{n,p})-p\right|
 =\Delta_{n,M}.
\end{equation}
Thus our distributional lower bound has an exact interpretation as a
uniform one-sided calibration error when the variance is known.

For the fixed nonlinear example, define
\[
 N_\gamma^{\mathrm{or}}(\varepsilon)
 =\inf\left\{N\in\mathbb N:
           \sup_{n\ge N}\Delta_{n,M}\le\varepsilon\right\},
\]
where the infimum of the empty set is infinity.
By \Cref{thm:nonlinear},
\begin{align}
 \Delta_{n,M}
 &\asymp n^{-(\gamma-1/2)}+n^{-(1-\gamma)},\nonumber\\
 N_\gamma^{\mathrm{or}}(\varepsilon)
 &\asymp
 \varepsilon^{-1/\min\{\gamma-1/2,\,1-\gamma\}}
 \qquad(\varepsilon\downarrow0).
 \label{eq:pr-complexity}
\end{align}
The smallest exponent in this oracle comparison is four, attained at
$\gamma=3/4$. Covariance estimation introduces a further requirement.

\subsection{Estimating the limiting variance from one trajectory}
The covariance lower bound uses an observation model that restricts
the estimator to a single SGD trajectory and its known step sizes.

\begin{assumption}
\label{ass:trajectory}
The estimator observes a single trajectory
$(\theta_0,\theta_1,\ldots,\theta_n)$ and the deterministic step sizes
$\alpha_k=1/[4(k+1)^\gamma]$. It has no access to stochastic Hessians,
raw observations, or additional oracle evaluations. The estimator
$\widehat\Sigma_n$ is a measurable function of this trajectory and
the known step sizes.
\end{assumption}
The stochastic gradients used along the trajectory can be recovered as
$(\theta_{k-1}-\theta_k)/\alpha_k$. Including these gradients among
the observations therefore does not change \Cref{ass:trajectory}.
In contrast, observing a stochastic Hessian can reveal information
that the trajectory does not contain.

Consider the following scalar quadratic family:
\begin{align}
 f_h(x)&=\frac h2x^2,
 & F_h(x,\xi)&=hx+\xi,
 & 1\le h&\le\frac32,\nonumber\\
 \theta_k&=(1-h\alpha_k)\theta_{k-1}-\alpha_k\xi_k,
 & \xi_k&\overset{\mathrm{i.i.d.}}\sim N(0,1),
 & \theta_0&=0.
 \label{eq:quadratic-family}
\end{align}
The unknown parameter is $h$. The optimum is zero and may be regarded
as known for this covariance-estimation problem. All models in this
family satisfy \Cref{ass:objective,ass:oracle,ass:polynomial} with the
same bounds $\mu=1/2$, $L_1=L_2=3/2$, $L_3=1/2$, noise variance one,
and fourth moment at most three. The nonlinear example satisfies
these same bounds. In the quadratic family,
\begin{equation}\label{eq:quadratic-target}
 \Sigma_\infty(h)=h^{-2}.
\end{equation}

\citet[Theorem~5.3]{NiHuo2026} obtain a covariance-estimation lower
bound by comparing nearby quadratic objectives through their
trajectory laws. Their theorem is stated in dimension at least two
and under a different step-size condition. We give the scalar
adaptation for the schedule used here, including a finite-sample
bound. The covariance-estimation obstruction is the one identified
in that work.

Write $P_h^{(n)}$ for the law of $(\theta_1,\ldots,\theta_n)$ under
\eqref{eq:quadratic-family}. Define the deterministic quantities
\begin{align}
 v_0&=0,
 &v_k&=(1-\alpha_k)^2v_{k-1}+\alpha_k^2,\nonumber\\
 \mathcal I_n&=\sum_{k=1}^n v_{k-1}.
 \label{eq:information-budget}
\end{align}
Here $v_k=\PE_1\theta_k^2$. The quantity $\mathcal I_n$ measures
the information in the trajectory about a perturbation of $h=1$.

The next lemma makes this interpretation precise by expressing the
Kullback--Leibler divergence between nearby trajectory laws in terms
of $\mathcal I_n$ and bounding its growth.

\begin{lemma}
\label{lem:trajectory-information}
For every $n\ge1$ and $0\le u\le1/2$,
\begin{equation}\label{eq:trajectory-kl}
 \operatorname{KL}\bigl(P_1^{(n)},P_{1+u}^{(n)}\bigr)
 =\frac{u^2}{2}\mathcal I_n.
\end{equation}
There exists a finite constant $D_\gamma$, depending only on $\gamma$,
such that
\begin{equation}\label{eq:information-growth}
 \mathcal I_n\le D_\gamma n^{1-\gamma}
 \qquad(n\ge1).
\end{equation}
\end{lemma}
\begin{proof}
Conditionally on $\theta_{k-1}=x$, the transition under parameter $h$
is the normal law
\[
 N\bigl((1-h\alpha_k)x,\alpha_k^2\bigr).
\]
The two transitions at $h=1$ and $h=1+u$ have the same variance,
and their means differ by $u\alpha_k x$. Their relative entropy is
therefore
\begin{align*}
 \operatorname{KL}\bigl(
 N((1-\alpha_k)x,\alpha_k^2),
 N((1-(1+u)\alpha_k)x,\alpha_k^2)\bigr) =\frac{(u\alpha_k x)^2}{2\alpha_k^2} =\frac{u^2x^2}{2}.
\end{align*}
The chain rule for relative entropy now gives
\begin{align*}
 \operatorname{KL}\bigl(P_1^{(n)},P_{1+u}^{(n)}\bigr) =\sum_{k=1}^n\PE_1\left[\frac{u^2\theta_{k-1}^2}{2}\right] =\frac{u^2}{2}\sum_{k=1}^n v_{k-1},
\end{align*}
which proves \eqref{eq:trajectory-kl}. To bound this sum, unroll the second-moment recursion in
\eqref{eq:information-budget}. Using \Cref{lem:kernel} with $b=2$
and $r=1$, we obtain
\begin{align*}
 v_k =\sum_{i=1}^k\alpha_i^2
       \prod_{\ell=i+1}^k(1-\alpha_\ell)^2 \le\sum_{i=1}^k\alpha_i^2
       \exp\left\{-2\sum_{\ell=i+1}^k\alpha_\ell\right\} \le C_{2,1}\alpha_k.
\end{align*}
Since $v_0=0$, it follows that
\begin{align*}
 \mathcal I_n \leq \frac{C_{2,1}}4\sum_{j=1}^{n-1}(j+1)^{-\gamma} \leq \frac{C_{2,1}}{4(1-\gamma)}n^{1-\gamma}.
\end{align*}
Thus one may take $D_\gamma=C_{2,1}/[4(1-\gamma)]$.
\end{proof}

The information calculation explains why convergence of the iterates
can make covariance estimation difficult. In this family, a
perturbation of the curvature changes the conditional gradient mean
by $u\theta_{k-1}$. Since $\PE_1\theta_{k-1}^2$ is of order
$\alpha_{k-1}$, the information about curvature grows only on the
scale $\sum_{k<n}\alpha_k$, rather than $n$. This is a concrete
instance of the decreasing information per query discussed by
\citet{RaginskyRakhlin2011}. Perturbations of size
$n^{-(1-\gamma)/2}$ consequently keep the trajectory divergence
bounded while changing $h^{-2}$ by the same order. The next theorem
turns this limitation into probability and mean-error lower bounds
for covariance estimation from one trajectory, using the two-point
construction of \citet{NiHuo2026}.

\begin{theorem}
\label{thm:covariance-lower}
Under \Cref{ass:trajectory}, every estimator $\widehat\Sigma_n$
satisfies, for every $n\ge1$,
\begin{align}
 \sup_{1\le h\le3/2}
 \PP_h\left(
   |\widehat\Sigma_n-h^{-2}|
      \ge\frac1{4\sqrt{1+\mathcal I_n}}\right)
 &\ge\frac38,\label{eq:covariance-probability-lower}\\
 \sup_{1\le h\le3/2}
 \PE_h|\widehat\Sigma_n-h^{-2}|
 &\ge\frac3{32\sqrt{1+\mathcal I_n}}.
 \label{eq:covariance-mean-lower}
\end{align}
Both inequalities remain valid when the absolute error is replaced
by the relative error $|\widehat\Sigma_n/h^{-2}-1|$. In particular,
there exists $c_\gamma>0$ such that
\begin{equation}\label{eq:covariance-rate-lower}
 \inf_{\widehat\Sigma_n}\sup_{1\le h\le3/2}
 \PE_h\left|\frac{\widehat\Sigma_n}{h^{-2}}-1\right|
 \ge c_\gamma n^{-(1-\gamma)/2}
 \qquad(n\ge1).
\end{equation}
\end{theorem}
\begin{proof}
Fix $n$ and choose two parameter values
\[
 h_0=1,\qquad h_1=1+u_n,\qquad
 u_n=\frac1{2\sqrt{1+\mathcal I_n}}.
\]
Since $0<u_n\le1/2$, both belong to the family
\eqref{eq:quadratic-family}. By \Cref{lem:trajectory-information},
\begin{align*}
 \operatorname{KL}\bigl(P_{h_0}^{(n)},P_{h_1}^{(n)}\bigr)
 &=\frac{\mathcal I_n}{8(1+\mathcal I_n)}
 \le\frac18.
\end{align*}
Pinsker's inequality implies
\begin{equation}\label{eq:trajectory-tv}
 \operatorname{TV}\bigl(P_{h_0}^{(n)},P_{h_1}^{(n)}\bigr)
 \le\frac14,
\end{equation}
where total variation is defined as the supremum of the difference
in probabilities over measurable events. The two target variances are separated by
\begin{align*}
 d_n =h_0^{-2}-h_1^{-2} =1-(1+u_n)^{-2} =u_n\frac{2+u_n}{(1+u_n)^2}
 \ge u_n.
\end{align*}
The last inequality holds for $0\le u_n\le1/2$.
Given $\widehat\Sigma_n$, select the parameter whose target variance
is closer to its value, breaking a tie deterministically. If this
test selects the wrong parameter under $h_j$, then
\[
 |\widehat\Sigma_n-h_j^{-2}|
 \ge d_n/2\ge u_n/2.
\]
The sum of the two testing error probabilities is at least
$1-\operatorname{TV}(P_{h_0}^{(n)},P_{h_1}^{(n)})$. Hence
\begin{align*}
 \max_{j\in\{0,1\}}
 \PP_{h_j}\left(
    |\widehat\Sigma_n-h_j^{-2}|\ge u_n/2\right) \ge\frac{1-\operatorname{TV}(P_{h_0}^{(n)},P_{h_1}^{(n)})}{2} \ge\frac38.
\end{align*}
This proves \eqref{eq:covariance-probability-lower}. Multiplying the
probability lower bound by $u_n/2$ proves
\eqref{eq:covariance-mean-lower}.

For every $h\in[1,3/2]$, we have $h^{-2}\le1$, and therefore
\[
 \left|\frac{\widehat\Sigma_n}{h^{-2}}-1\right|
 =h^2|\widehat\Sigma_n-h^{-2}|
 \ge|\widehat\Sigma_n-h^{-2}|.
\]
This proves the relative-error assertions. Finally,
\eqref{eq:information-growth} gives
\[
 \frac3{32\sqrt{1+\mathcal I_n}}
 \ge\frac3{32\sqrt{1+D_\gamma}}n^{-(1-\gamma)/2},
\]
which yields \eqref{eq:covariance-rate-lower}.
\end{proof}

The supremum over $h$ is essential: the two alternatives in this
proof depend on $n$. This is a uniform estimation lower bound, in
contrast to \Cref{thm:nonlinear}, which uses one fixed nonlinear model.
It already holds when the optimum and the noise variance are known;
the unknown curvature determines the difficulty of estimating
$\Sigma_\infty$. 

\subsection{Joint accuracy and necessary sample size}
Let $\mathcal M_\gamma$ consist of the nonlinear example
\eqref{eq:nonlinear-example} and the quadratic family
\eqref{eq:quadratic-family}, all with the fixed schedule
$\alpha_k=1/[4(k+1)^\gamma]$. For an estimator satisfying
\Cref{ass:trajectory}, define the joint risk
\begin{equation}\label{eq:joint-risk}
 \mathfrak R_n(\gamma)
 =\inf_{\widehat\Sigma_n}\sup_{M\in\mathcal M_\gamma}
 \max\left\{
   \Delta_{n,M},\;
   \PE_M\left|\frac{\widehat\Sigma_n}{\Sigma_\infty(M)}-1\right|
 \right\}.
\end{equation}
This criterion requires both an accurate Gaussian approximation with
the limiting variance and an accurate estimate of that variance.
The same estimator must be used throughout the model class. Every
larger class containing $\mathcal M_\gamma$ inherits the lower bound
below, provided the observation restriction is unchanged.

\Needspace{6\baselineskip}
Combining the Gaussian approximation and covariance-estimation
obstructions yields the following lower bound on the joint risk and
its associated iteration complexity.

\begin{theorem}
\label{thm:inference-complexity}
For every fixed $\gamma\in(1/2,1)$, there exist constants
$a_\gamma>0$ and $n_\gamma\ge1$ such that
\begin{equation}\label{eq:joint-risk-lower}
 \mathfrak R_n(\gamma)
 \ge a_\gamma\left\{
    n^{-(\gamma-1/2)}+n^{-(1-\gamma)/2}
 \right\}
 \qquad(n\ge n_\gamma).
\end{equation}
Define
\begin{equation}\label{eq:iteration-complexity}
 \mathfrak N_\gamma(\varepsilon)
 =\inf\left\{N\in\mathbb N:
       \sup_{n\ge N}\mathfrak R_n(\gamma)\le\varepsilon\right\}.
\end{equation}
For all sufficiently small $\varepsilon>0$,
\begin{equation}\label{eq:joint-complexity-lower}
 \mathfrak N_\gamma(\varepsilon)
 \ge a_\gamma'
 \varepsilon^{-\max\{1/(\gamma-1/2),\,2/(1-\gamma)\}},
\end{equation}
with $a_\gamma'>0$. The smallest exponent furnished by
\eqref{eq:joint-complexity-lower} is six, at $\gamma=2/3$.
\end{theorem}
\begin{proof}
Let $M_0$ denote the fixed nonlinear model. The first component in
\eqref{eq:joint-risk} does not depend on the covariance estimator.
By \Cref{thm:nonlinear}, for all sufficiently large $n$,
\begin{equation}\label{eq:joint-first-obstruction}
 \mathfrak R_n(\gamma)
 \ge\Delta_{n,M_0}
 \ge b_\gamma\left\{
        n^{-(\gamma-1/2)}+n^{-(1-\gamma)}\right\}
\end{equation}
for some $b_\gamma>0$. On the quadratic subfamily,
\Cref{thm:covariance-lower} gives
\begin{equation}\label{eq:joint-second-obstruction}
 \mathfrak R_n(\gamma)
 \ge\inf_{\widehat\Sigma_n}\sup_{1\le h\le3/2}
      \PE_h\left|\frac{\widehat\Sigma_n}{h^{-2}}-1\right|
 \ge c_\gamma n^{-(1-\gamma)/2}.
\end{equation}
Taking the maximum of \eqref{eq:joint-first-obstruction} and
\eqref{eq:joint-second-obstruction}, and using
$\max\{x,y\}\ge(x+y)/2$ for $x,y\ge0$, proves
\eqref{eq:joint-risk-lower}. The variance-estimation term is larger
than $n^{-(1-\gamma)}$, which explains why the latter does not
appear in the simplified bound.

Put
\[
 q_\gamma=\min\{\gamma-1/2,(1-\gamma)/2\}.
\]
Then \eqref{eq:joint-risk-lower} implies
$\mathfrak R_n(\gamma)\ge a_\gamma n^{-q_\gamma}$ for
$n\ge n_\gamma$. For sufficiently small $\varepsilon$, every finite
$N$ in the set defining \eqref{eq:iteration-complexity} exceeds
$n_\gamma$. Applying the last bound at $n=N$ gives
\[
 N\ge(a_\gamma/\varepsilon)^{1/q_\gamma},
\]
and proves \eqref{eq:joint-complexity-lower}. The function
$\gamma-1/2$ is increasing, whereas $(1-\gamma)/2$ is decreasing.
They coincide when
\[
 \gamma-\frac12=\frac{1-\gamma}{2},
 \qquad\text{that is,}\qquad\gamma=\frac23.
\]
At this value $q_\gamma=1/6$, completing the proof.
\end{proof}

The two contributions in \Cref{thm:inference-complexity} arise on
different members of the model class. The theorem concerns the joint
requirements in \eqref{eq:joint-risk}; it is not a lower bound for
the coverage error of every confidence-interval procedure. In
particular, a lower bound for estimating a variance does not by itself
give a lower bound for coverage after studentization. Nor does the
necessary exponent six establish a matching upper bound or optimality
of $\gamma=2/3$ for a particular feasible method.

\Needspace{10\baselineskip}
\subsection{Relation to feasible Gaussian intervals}
For completeness, the following bound explains how the two errors
enter a sufficient coverage guarantee. It permits estimation of the
variance from the same trajectory as the average.

\begin{lemma}
\label{lem:studentization-transfer}
Fix a model $M$, write $\Sigma=\Sigma_\infty(M)$, and suppose that
$\widehat\Sigma_n>0$ almost surely. Set
\[
 T_n=\frac{\sqrt n(\prtheta_n-\theta_M^\star)}
                {\sqrt{\widehat\Sigma_n}},
 \qquad
 r_{n,M}=\PE_M\left|\frac{\widehat\Sigma_n}{\Sigma}-1\right|.
\]
There is a numerical constant $C$ such that, for every
$0<\eta\le1/2$,
\begin{align}
 \dk(T_n,N(0,1))
 &\le\Delta_{n,M}+C\eta
       +\PP_M\left(
          \left|\frac{\widehat\Sigma_n}{\Sigma}-1\right|>\eta
         \right)\nonumber\\
 &\le\Delta_{n,M}+C\eta+\frac{r_{n,M}}\eta.
 \label{eq:studentization-transfer}
\end{align}
\end{lemma}
\begin{proof}
Let $Y_n=\sqrt n(\prtheta_n-\theta_M^\star)/\sqrt\Sigma$ and
$E_\eta=\{|\widehat\Sigma_n/\Sigma-1|\le\eta\}$. For a fixed
$z\in\rset$, define
\begin{align*}
 a_z&=\min\{z\sqrt{1-\eta},z\sqrt{1+\eta}\},\\
 b_z&=\max\{z\sqrt{1-\eta},z\sqrt{1+\eta}\}.
\end{align*}
On $E_\eta$, the random threshold
$z\sqrt{\widehat\Sigma_n/\Sigma}$ lies in $[a_z,b_z]$. Consequently,
\begin{align*}
 \PP_M(Y_n\le a_z)-\PP_M(E_\eta^c)
 &\le\PP_M(T_n\le z)\\
 &\le\PP_M(Y_n\le b_z)+\PP_M(E_\eta^c).
\end{align*}
No independence is used in these inclusions. By
\eqref{eq:calibration-error}, either distribution function of $Y_n$
differs from the corresponding standard normal probability by at
most $\Delta_{n,M}$. Moreover, for
$a\in[1/\sqrt2,\sqrt{3/2}]$,
\[
 \sup_{z\in\rset}
 \left|\frac{\partial}{\partial a}\Phi(az)\right|
 =\sup_{z\in\rset}|z|\ph(az)
 \le\sqrt2\sup_{x\in\rset}|x|\ph(x).
\]
Since $|\sqrt{1\pm\eta}-1|\le\eta$ for $0<\eta\le1/2$,
the normal probabilities at $a_z$ and $b_z$ differ from $\Phi(z)$
by at most $C\eta$, uniformly in $z$. This proves the first line of
\eqref{eq:studentization-transfer}. Markov's inequality gives the
second line.
\end{proof}

Choosing $\eta$ of order $\sqrt{r_{n,M}}$ gives the general sufficient
bound $\Delta_{n,M}+O(\sqrt{r_{n,M}})$ when $r_{n,M}$ is small.
Sharper coverage bounds may use additional properties of the joint
law of the average and its variance estimator. This distinction is
needed when comparing the joint complexity in
\Cref{thm:inference-complexity} with the performance of plug-in,
batch-means, or bootstrap intervals
\citep{chen2020,ZhuChenWu2023,WeiZhuLyuWu2026,sgd}.

\section*{Acknowledgments}
This work is an output of a research project HSE-BR-2025-019 implemented as part of the Basic Research Program at HSE University.

\section*{AI usage statement}
The authors used ChatGPT (OpenAI) to assist with developing and checking mathematical arguments and with drafting and revising the manuscript. The authors take full responsibility for the content of the paper, including the correctness of all results, proofs, and references.

\bibliographystyle{plainnat}
\bibliography{sgd_lower_bound}

\clearpage
\appendix
\section{Expansion}

\subsection{The linearization and its corrections}\label{sec:expansion}
The proof starts from an exact decomposition of the recursion. We give
its construction before estimating its terms. This also explains how
the perturbation expansion for linear stochastic approximation in
\citet{durmus} enters the nonlinear problem.

Write $q(x)=1-\cos x$. The gradient of our objective satisfies
\begin{align}
 f'(x)&=x+\frac12q(x),\label{eq:gradient-split}\\
 \left|q(x)-\frac{x^2}{2}\right|&\le\frac{x^4}{24},
 \qquad 0\le q(x)\le\frac{x^2}{2}.
 \label{eq:q-taylor}
\end{align}
Consequently, the exact SGD update can be written as
\begin{align}
 \theta_k
 &=(1-\alpha_k)\theta_{k-1}-\alpha_k S_k
       -\alpha_kR_k\theta_{k-1}\notag\\
 &\hspace{18mm}-\frac{\alpha_k}{2}(1+R_k)q(\theta_{k-1}).
 \label{eq:theta-expanded}
\end{align}
The first two terms in \eqref{eq:theta-expanded} define the principal
linear statistic. We therefore set $U_0=0$ and define
\begin{equation}
 U_k=(1-\alpha_k)U_{k-1}-\alpha_kS_k.
 \label{eq:nonlinear-U}
\end{equation}
The process $U$ retains the additive noise and replaces the random
linear coefficient $1+R_k$ by its mean, which is one.

\paragraph{Connection with the linear perturbation expansion.}
For comparison, consider the linear recursion
\[
 L_k=\{1-\alpha_k(1+R_k)\}L_{k-1}-\alpha_kS_k,
 \qquad L_0=0.
\]
In the notation of \citet{durmus}, its centered random coefficient is
$\widetilde A_k=R_k$, its mean coefficient is $\bar A=1$, and its noise
at the solution is $\varepsilon_k=S_k$. Thus its principal term
$J_k^{(0)}$ is exactly $U_k$. Subtracting the recursion for $U_k$
from that for $L_k$ gives
\begin{align}
 H_{k,\mathrm{lin}}^{(0)}
   &:=L_k-U_k,\notag\\
 H_{k,\mathrm{lin}}^{(0)}
   &=\{1-\alpha_k(1+R_k)\}H_{k-1,\mathrm{lin}}^{(0)}
       -\alpha_kR_kU_{k-1}.
 \label{eq:classical-H0}
\end{align}
The next step in their expansion replaces the random coefficient
multiplying this remainder by its mean. This gives $J_0=0$ and
\begin{equation}
 J_k=(1-\alpha_k)J_{k-1}-\alpha_kR_kU_{k-1}.
 \label{eq:nonlinear-J}
\end{equation}
Hence our $J_k$ is their $J_k^{(1)}$. In the linear recursion,
subtracting \eqref{eq:nonlinear-J} from \eqref{eq:classical-H0}
would give the next remainder
\begin{align*}
 H_{k,\mathrm{lin}}^{(1)}
   &:=L_k-U_k-J_k,\\
 H_{k,\mathrm{lin}}^{(1)}
   &=\{1-\alpha_k(1+R_k)\}H_{k-1,\mathrm{lin}}^{(1)}
       -\alpha_kR_kJ_{k-1}.
\end{align*}
These are the $J^{(0)}$--$H^{(0)}$ and $J^{(1)}$--$H^{(1)}$
recursions of \citet{durmus}, here written with the possibly varying
step size $\alpha_k$.

\paragraph{The additional correction caused by curvature.}
In the nonlinear recursion, $e_k:=\theta_k-U_k$ instead satisfies
\begin{align}
 e_k = \{1-\alpha_k(1+R_k)\}e_{k-1} -\alpha_kR_kU_{k-1} -\frac{\alpha_k}{2}(1+R_k)q(\theta_{k-1}).
 \label{eq:nonlinear-H0}
\end{align}
The last term is absent from \eqref{eq:classical-H0}. Its deterministic
quadratic part contributes to the leading distributional error.
Indeed, \eqref{eq:q-taylor} gives
\[
 \frac12q(x)=\frac{x^2}{4}+r(x),
 \qquad |r(x)|\le\frac{x^4}{48}.
\]
To retain this contribution, replace $\theta_{k-1}$ by $U_{k-1}$ in
that quadratic part, define $K_0=0$, and
\begin{equation}
 K_k=(1-\alpha_k)K_{k-1}-\frac{\alpha_k}{4}U_{k-1}^2.
 \label{eq:nonlinear-K}
\end{equation}
Unlike $J_k$, the correction $K_k$ has a nonzero mean. Its negative
sign will determine the direction of the leading bias. The remaining terms are collected in
\begin{equation}
 \theta_k=U_k+J_k+K_k+H_k,
 \qquad H_0=0.
 \label{eq:full-expansion}
\end{equation}
This is an exact identity. To derive the recursion for $H_k$, first
rewrite \eqref{eq:theta-expanded} as
\[
 \theta_k=(1-\alpha_k)\theta_{k-1}-\alpha_kS_k
       -\frac{\alpha_k}{2}q(\theta_{k-1})
       -\alpha_kR_k f'(\theta_{k-1}).
\]
Subtracting \eqref{eq:nonlinear-U}, \eqref{eq:nonlinear-J}, and
\eqref{eq:nonlinear-K} then gives
\begin{align}
 H_k
 &=(1-\alpha_k)H_{k-1}
   -\frac{\alpha_k}{2}
      \left\{q(\theta_{k-1})-\frac{U_{k-1}^2}{2}\right\}
      \notag\\
 &\hspace{18mm}
   -\alpha_kR_k\{f'(\theta_{k-1})-U_{k-1}\}.
 \label{eq:nonlinear-H}
\end{align}
Our $H_k$ therefore includes the nonlinear terms left after removing
$K_k$; it is not the process $H_{k,\mathrm{lin}}^{(1)}$ above.

\paragraph{Deterministic propagation of the forcing terms.}
We use the weights $P_{i,k}$ and $Q_{i,n}$ defined in \eqref{eq:weights}. Iterating each recursion gives
\begin{align}
 U_k&=-\sum_{i=1}^k\alpha_iP_{i,k}S_i,\notag\\
 J_k&=-\sum_{i=1}^k\alpha_iP_{i,k}R_iU_{i-1},\notag\\
 K_k&=-\frac14\sum_{i=1}^k\alpha_iP_{i,k}U_{i-1}^2.
 \label{eq:correction-sums}
\end{align}
For every fixed terminal index $k$, all the weights $P_{i,k}$ are
deterministic. With $\filtr_i=\sigma(R_j,S_j:1\le j\le i)$, the
variables $R_iU_{i-1}$ and
$R_i\{f'(\theta_{i-1})-U_{i-1}\}$ are $\filtr_i$-measurable and
have conditional mean zero given $\filtr_{i-1}$. This is the
martingale difference property used below. In particular, no
martingale property is assigned to a sum containing future random
propagators.

The estimates in \Cref{sec:preliminaries} show that $U_k$ has
$L^2$ norm of order at most $\sqrt{\alpha_k}$, whereas $J_k$ and
$K_k$ have $L^2$ norm of order at most $\alpha_k$. The final remainder
has $L^1$ norm of order at most $\alpha_k^{3/2}$. We will retain the
signed effect of $K_k$ in the test expectation; a bound on its norm
alone would lose the contribution needed for the lower bound.

\subsection{Averaging the perturbation expansion}
We first average the decomposition
$\theta_k=U_k+J_k+K_k+H_k$. The deterministic weights in
\eqref{eq:weights} make it possible to separate the mean curvature
correction from the centered terms.

We first bound the normalized average of the remainder $H_k$, whose
contribution will be negligible at the scale of the curvature bias.

\begin{lemma}\label{lem:pr-remainder}
For the remainder in \eqref{eq:nonlinear-H},
\begin{equation}
 \left\|\frac1{\sqrt n}\sum_{k=0}^{n-1}H_k\right\|_1
 \le \frac C{\sqrt n}
       \left(1+\sum_{i=1}^{n-1}\alpha_i^{3/2}\right).
 \label{eq:nonlinear-H-bound}
\end{equation}
\end{lemma}
\begin{proof}
For $i\ge1$, put
\begin{align*}
 \Delta_i&=q(\theta_{i-1})-\frac12U_{i-1}^2,\\
 \Xi_i&=R_i\{f'(\theta_{i-1})-U_{i-1}\}.
\end{align*}
Since $H_0=0$, unrolling \eqref{eq:nonlinear-H} gives
\begin{align*}
 H_k
 &=-\sum_{i=1}^k\alpha_iP_{i,k}
       \left(\frac12\Delta_i+\Xi_i\right),\\
 \frac1{\sqrt n}\sum_{k=0}^{n-1}H_k
 &=-\frac1{2\sqrt n}\sum_{i=1}^{n-1}Q_{i,n}\Delta_i
   -\frac1{\sqrt n}\sum_{i=1}^{n-1}Q_{i,n}\Xi_i.
\end{align*}
The second equality follows by interchanging two finite sums.
By \Cref{lem:kernel,lem:forcing}, the first sum satisfies
\begin{align*}
 \left\|\frac1{2\sqrt n}
        \sum_{i=1}^{n-1}Q_{i,n}\Delta_i\right\|_1 \leq \frac1{2\sqrt n}
       \sum_{i=1}^{n-1}Q_{i,n}\PE|\Delta_i| \leq \frac C{\sqrt n}
       \sum_{i=1}^{n-1}\alpha_i^{3/2}.
\end{align*}
For the second sum, let
$\filtr_i=\sigma(R_j,S_j:1\le j\le i)$. The variable $\Xi_i$ is
$\filtr_i$-measurable and
$\PE[\Xi_i\mid\filtr_{i-1}]=0$. Thus distinct summands are
orthogonal. Using the second forcing estimate in
\Cref{lem:forcing}, we obtain
\begin{align*}
 \left\|\frac1{\sqrt n}
        \sum_{i=1}^{n-1}Q_{i,n}\Xi_i\right\|_2^2 =\frac1n \sum_{i=1}^{n-1}Q_{i,n}^2\PE\Xi_i^2 \leq \frac Cn\sum_{i=1}^{n-1}\alpha_i^2
 \le\frac Cn.
\end{align*}
The last series is bounded because $2\gamma>1$. The triangle
inequality and $\|V\|_1\le\|V\|_2$ now prove the claim.
\end{proof}

We next control the fluctuations of the multiplicative correction
and of the centered curvature correction after averaging.

\begin{lemma}
\label{lem:pr-centered}
The corrections in \eqref{eq:nonlinear-J} and
\eqref{eq:nonlinear-K} satisfy
\begin{align}
 \left\|\frac1{\sqrt n}\sum_{k=0}^{n-1}J_k\right\|_2^2
 &\le Cn^{-\gamma},\label{eq:pr-centered-J}\\
 \left\|\frac1{\sqrt n}\sum_{k=0}^{n-1}
                 (K_k-\PE K_k)\right\|_2^2
 &\le Cn^{-\gamma}.\label{eq:pr-centered-K}
\end{align}
\end{lemma}
\begin{proof}
Write $v_k=\PE U_k^2$, so $v_0=0$. Unrolling and averaging
\eqref{eq:nonlinear-J} yields
\[
 \frac1{\sqrt n}\sum_{k=0}^{n-1}J_k
 =-\frac1{\sqrt n}\sum_{i=1}^{n-1}Q_{i,n}R_iU_{i-1}.
\]
The sequence $R_iU_{i-1}$ consists of martingale differences for
$(\filtr_i)$, and $\PE R_i^2=1$. Consequently,
\begin{align*}
 \left\|\frac1{\sqrt n}\sum_{k=0}^{n-1}J_k\right\|_2^2
 &=\frac1n\sum_{i=1}^{n-1}Q_{i,n}^2v_{i-1}\\
 &\le\frac Cn\sum_{i=2}^{n-1}\alpha_{i-1}
 \le Cn^{-\gamma}.
\end{align*}
This proves \eqref{eq:pr-centered-J}.

The curvature correction is not a sum of martingale differences. Its centered average has the exact representation
\[
 \frac1{\sqrt n}\sum_{k=0}^{n-1}(K_k-\PE K_k)
 =-\frac1{4\sqrt n}\sum_{i=1}^{n-1}
       Q_{i,n}(U_{i-1}^2-v_{i-1}).
\]
We therefore compute its covariance directly. If $1\le a\le b$, then
\begin{align*}
 U_b&=P_{a,b}U_a+V_{a,b},\\
 V_{a,b}&=-\sum_{\ell=a+1}^b\alpha_\ell P_{\ell,b}S_\ell.
\end{align*}
The random variable $V_{a,b}$ is centered and independent of $U_a$.
Expanding the square and taking expectations gives
\begin{align*}
 \PE[U_a^2U_b^2]
 &=P_{a,b}^2\PE U_a^4
   +2P_{a,b}\PE U_a^3\,\PE V_{a,b}
   +\PE U_a^2\,\PE V_{a,b}^2\\
 &=P_{a,b}^2\PE U_a^4+v_a\PE V_{a,b}^2,\\
 v_av_b
 &=P_{a,b}^2v_a^2+v_a\PE V_{a,b}^2.
\end{align*}
Subtracting the last two identities proves
\begin{equation}
 \Cov(U_a^2,U_b^2)
 =P_{a,b}^2\var(U_a^2).
 \label{eq:pr-square-covariance}
\end{equation}
In particular, for $1\le a\le b$, \Cref{lem:moments} implies
\[
 0\le\Cov(U_a^2,U_b^2)
 \le C P_{a,b}^2\alpha_a^2.
\]
Terms with $a=0$ vanish because $U_0=0$.

We now sum this bound. By \Cref{lem:kernel},
\begin{align*}
 \sum_{j=i}^{n-1}P_{i-1,j-1}^2
 &\le\sum_{b=i-1}^{\infty}
      \exp\left(-2\sum_{\ell=i}^{b}\alpha_\ell\right)
 \le\frac C{\alpha_{i-1}},\qquad i\ge2.
\end{align*}
Since $Q_{i,n}$ is nonnegative and uniformly bounded, it follows that
\begin{align*}
 \left\|\frac1{\sqrt n}\sum_{k=0}^{n-1}
                    (K_k-\PE K_k)\right\|_2^2 &= \frac1{16n}\sum_{i=1}^{n-1}\sum_{j=1}^{n-1}
        Q_{i,n}Q_{j,n}\Cov(U_{i-1}^2,U_{j-1}^2)\\
 &\quad\le\frac Cn\sum_{i=2}^{n-1}
           \alpha_{i-1}^2\sum_{j=i}^{n-1}P_{i-1,j-1}^2\\
 &\quad\le\frac Cn\sum_{i=2}^{n-1}\alpha_{i-1}
 \le Cn^{-\gamma}.
\end{align*}
In the first inequality we used the symmetry of the double sum and
bounded it by twice the sum over $i\le j$. This proves
\eqref{eq:pr-centered-K}.
\end{proof}

The mean curvature correction produces a deterministic shift in the
normalized average. The next lemma identifies its leading coefficient.

\begin{lemma}\label{lem:pr-bias}
Define the nonnegative deterministic quantity
\begin{equation}
 B_n=\frac1{4\sqrt n}\sum_{i=1}^{n-1}Q_{i,n}v_{i-1}.
 \label{eq:pr-bias-definition}
\end{equation}
Then
\begin{equation}
 B_n\sim\frac{n^{1/2-\gamma}}{32(1-\gamma)}.
 \label{eq:nonlinear-bias}
\end{equation}
\end{lemma}
\begin{proof}
The linear recursion \eqref{eq:nonlinear-U} and the independence of
$S_k$ imply
\begin{align*}
 v_k&=(1-\alpha_k)^2v_{k-1}+\alpha_k^2,\\
 \frac{v_k}{\alpha_k}
 &=(1-\alpha_k)^2\frac{\alpha_{k-1}}{\alpha_k}
      \frac{v_{k-1}}{\alpha_{k-1}}+\alpha_k.
\end{align*}
For the polynomial step sizes,
\[
 \frac{\alpha_{k-1}}{\alpha_k}
 =\left(1+\frac1k\right)^\gamma
 =1+O(k^{-1})=1+o(\alpha_k).
\]
Thus the coefficient multiplying $v_{k-1}/\alpha_{k-1}$ is
$1-2\alpha_k+o(\alpha_k)$. Applying \Cref{lem:tracking} gives
\begin{equation}
 \frac{v_k}{\alpha_k}\longrightarrow\frac12.
 \label{eq:pr-v-limit}
\end{equation}

Next, define $t_0=0$ and
\[
 t_k=\sum_{i=1}^k\alpha_iP_{i,k}v_{i-1},\qquad k\ge1.
\]
The deterministic convolution satisfies
\begin{align*}
 t_k&=(1-\alpha_k)t_{k-1}+\alpha_kv_{k-1},\\
 \frac{t_k}{\alpha_k}
 &=(1-\alpha_k)\frac{\alpha_{k-1}}{\alpha_k}
      \frac{t_{k-1}}{\alpha_{k-1}}+v_{k-1}\\
 &=\{1-\alpha_k+o(\alpha_k)\}
      \frac{t_{k-1}}{\alpha_{k-1}}
      +\alpha_k\{1/2+o(1)\}.
\end{align*}
We used \eqref{eq:pr-v-limit} and
$\alpha_{k-1}/\alpha_k\to1$ in the last equality. A second application
of \Cref{lem:tracking} yields $t_k/\alpha_k\to1/2$.

Finally, interchanging sums in \eqref{eq:pr-bias-definition},
\begin{align*}
 4\sqrt n B_n =\sum_{i=1}^{n-1}\alpha_iv_{i-1}
                  \sum_{k=i}^{n-1}P_{i,k} =\sum_{k=1}^{n-1}\sum_{i=1}^k
                  \alpha_iP_{i,k}v_{i-1}
 =\sum_{k=1}^{n-1}t_k.
\end{align*}
Since $\sum_k\alpha_k=\infty$, the limit $t_k/\alpha_k\to1/2$
implies
\begin{align*}
 \sum_{k=1}^{n-1}t_k \sim\frac12\sum_{k=1}^{n-1}\alpha_k
 =\frac18\sum_{k=1}^{n-1}(k+1)^{-\gamma} \sim\frac{n^{1-\gamma}}{8(1-\gamma)}\,.
\end{align*}
The last equivalence follows by integral comparison. Dividing by
$4\sqrt n$ proves \eqref{eq:nonlinear-bias}.
\end{proof}

Combining these estimates reduces the normalized average to a
weighted sum of independent noise variables and a deterministic shift,
up to a smaller remainder.

\begin{lemma}\label{lem:pr-reduction}
Recall
\[
 W_n=-\frac1{\sqrt n}\sum_{i=1}^{n-1}Q_{i,n}S_i.
\]
Then
\begin{equation}
 X_n=W_n-B_n+\mathcal R_n,
 \label{eq:nonlinear-final-expansion}
\end{equation}
where $B_n$ is defined in \eqref{eq:pr-bias-definition} and
\begin{align}
 \|\mathcal R_n\|_1
 &\le Cn^{-\gamma/2}
       +\frac C{\sqrt n}
        \left(1+\sum_{i=1}^{n-1}\alpha_i^{3/2}\right)
   =o(n^{1/2-\gamma}).\label{eq:nonlinear-small-remainder}
\end{align}
Moreover,
\begin{equation}
 \dk(W_n,Z_n)\le Cn^{-1/2}.
 \label{eq:pr-weighted-be}
\end{equation}
\end{lemma}
\begin{proof}
Unrolling \eqref{eq:nonlinear-U} shows that
$n^{-1/2}\sum_{k=0}^{n-1}U_k=W_n$. Likewise,
$n^{-1/2}\sum_{k=0}^{n-1}\PE K_k=-B_n$. Therefore
\eqref{eq:nonlinear-final-expansion} holds with
\[
 \mathcal R_n=
 \frac1{\sqrt n}\sum_{k=0}^{n-1}
       \{J_k+(K_k-\PE K_k)+H_k\}.
\]
The norm estimate follows from
\Cref{lem:pr-remainder,lem:pr-centered}. To verify that it is a
little-oh bound, first note that
\[
 \frac{n^{-\gamma/2}}{n^{1/2-\gamma}}
 =n^{(\gamma-1)/2}\longrightarrow0.
\]
For the remaining term, integral comparison gives
\[
 1+\sum_{i=1}^{n-1}\alpha_i^{3/2}
 \le C\begin{cases}
 n^{1-3\gamma/2},&1/2<\gamma<2/3,\\
 1+\log n,&\gamma=2/3,\\
 1,&2/3<\gamma<1.
 \end{cases}
\]
After multiplication by $n^{\gamma-1}$, each right-hand side tends
to zero. This proves \eqref{eq:nonlinear-small-remainder}.

Finally, the summands defining $W_n$ are independent and centered,
and $\var(W_n)=s_n^2$. By \Cref{lem:covariance}, $s_n^2\to1$.
The Berry--Esseen inequality for independent sums, or the linear
case of \citet[Theorem~2.1]{shao2022}, therefore gives
\begin{align*}
 \dk(W_n,Z_n)
 &\le \frac C{s_n^3n^{3/2}}
        \sum_{i=1}^{n-1}Q_{i,n}^3\PE|S_i|^3\\
 &\le Cn^{-1/2}.
\end{align*}
Here $\PE|S_i|^3=1$ and $Q_{i,n}\le C$ by \Cref{lem:kernel}.
\end{proof}

\section{Technical bounds}
\label{sec:preliminaries}
The estimates below apply to the fixed objective and oracle in
\eqref{eq:nonlinear-example}, always with zero initialization. We use
the fixed schedule in \eqref{eq:nonlinear-recursion}.
Constants may depend on $\gamma$ but not on
the iteration index. The deterministic kernel bounds depend only on
the step-size schedule. For this schedule,
$\alpha_{k-1}\le2\alpha_k$ when $k\ge2$.

\subsection{Deterministic kernel bounds}
We begin with bounds on the deterministic propagation kernels and
the averaging weights that appear in the perturbation expansion.

\begin{lemma}\label{lem:kernel}
For every $b>0$ and $r\ge0$, there are finite constants $C_{b,r}$
and $C_b$ such that
\begin{align}
 \sum_{i=1}^k\alpha_i^{1+r}
    \exp\left\{-b\sum_{\ell=i+1}^k\alpha_\ell\right\}
    &\le C_{b,r}\alpha_k^r,\notag\\
 \sum_{k=i}^{\infty}
    \exp\left\{-b\sum_{\ell=i+1}^k\alpha_\ell\right\}
    &\le C_b\alpha_i^{-1}.
 \label{eq:kernel}
\end{align}
In particular, the weights in \eqref{eq:weights} satisfy
$0\le Q_{i,n}\le C$.
\end{lemma}
\begin{proof}
Let $S_k$ denote the first sum in \eqref{eq:kernel}, and put
$z_k=S_k/\alpha_k^r$. For $k\ge2$,
\begin{align*}
 S_k&=e^{-b\alpha_k}S_{k-1}+\alpha_k^{1+r},\\
 z_k&=e^{-b\alpha_k}
       \left(\frac{\alpha_{k-1}}{\alpha_k}\right)^r z_{k-1}
       +\alpha_k.
\end{align*}
Since $(\alpha_{k-1}/\alpha_k)^r=(1+1/k)^{\gamma r}
=1+O(k^{-1})$ and $k^{-1}=o(\alpha_k)$, the coefficient of
$z_{k-1}$ is $1-b\alpha_k+o(\alpha_k)$.
The asymptotic consequence of \Cref{lem:tracking} gives $z_k\to1/b$.
Thus $(z_k)$ is bounded, proving the first estimate for every fixed
$r\ge0$.

For the forward sum, integral comparison gives
\[
 b\sum_{\ell=i+1}^k\alpha_\ell
 \ge\frac{b}{4(1-\gamma)}
       \bigl\{(k+2)^{1-\gamma}-(i+2)^{1-\gamma}\bigr\}.
\]
Apply \citet[Lemma~31]{SamsonovEtAl2025Markov} with
$A=b/[4(1-\gamma)]$ and starting index $i+2$.
Its bound is uniform in the upper endpoint, so monotone convergence
yields
\[
 \sum_{k=i}^{\infty}
   \exp\left\{-b\sum_{\ell=i+1}^k\alpha_\ell\right\}
 \le C_{b,\gamma}(i+2)^\gamma
 \le C_b\alpha_i^{-1}.
\]
Finally, $0\le P_{i,k}\le
\exp\{-\sum_{\ell=i+1}^k\alpha_\ell\}$, and hence
$0\le Q_{i,n}\le C$ by the forward estimate with $b=1$.
\end{proof}

The following deterministic estimate controls how a perturbed
contracting recursion approaches its equilibrium. It will be used to
obtain the asymptotics of the moment and covariance recursions.

\begin{lemma}\label{lem:tracking}
Let $0\le m<n$, $b>0$, and $d\in\rset$. Suppose that, for
$m<k\le n$,
\[
 z_k=(1-ba_k+a_k\rho_k)z_{k-1}+a_k(d+\eta_k),
 \qquad 0<a_k\le b^{-1},\quad |\rho_k|\le b/2.
\]
Then
\begin{equation}
\label{eq:tracking-uniform}
 \left|z_n-\frac db\right|
 \le \exp\left\{-\frac b2\sum_{k=m+1}^n a_k\right\}
       \left|z_m-\frac db\right|
       +\frac2b\max_{m<k\le n}
          \left|\eta_k+\frac db\rho_k\right|.
\end{equation}
In particular, if $a_k\to0$, $\sum_{k\ge1}a_k=\infty$, and
\[
 z_k=\{1-ba_k+o(a_k)\}z_{k-1}+a_k\{d+o(1)\},
\]
then $z_k\to d/b$.
\end{lemma}
\begin{proof}
Set $u_k=z_k-d/b$ and
$\delta=\max_{m<k\le n}|\eta_k+(d/b)\rho_k|$. Since
$1-ba_k\ge0$ and $|\rho_k|\le b/2$,
\begin{align*}
 u_k&=(1-ba_k+a_k\rho_k)u_{k-1}
       +a_k\left(\eta_k+\frac db\rho_k\right),\\
 |u_k|&\le\left(1-\frac b2a_k\right)|u_{k-1}|+a_k\delta.
\end{align*}
The number $2\delta/b$ is a fixed point of the affine map on the
right. Subtracting it and iterating therefore gives
\[
 |u_n|-\frac{2\delta}{b}
 \le\prod_{k=m+1}^n\left(1-\frac b2a_k\right)
       \left(|u_m|-\frac{2\delta}{b}\right).
\]
Dropping the nonpositive term on the right and using $1-x\le e^{-x}$
proves \eqref{eq:tracking-uniform}.

For the last assertion, write the residuals as $a_k\rho_k$ and
$a_k\eta_k$, where $\rho_k,\eta_k\to0$. Given $\varepsilon>0$,
choose $m$ so that the step-size and residual bounds hold and
$|\eta_k+(d/b)\rho_k|\le\varepsilon$ for every $k>m$.
As $n\to\infty$, the exponential term in
\eqref{eq:tracking-uniform} vanishes. Hence
\[
 \limsup_{n\to\infty}\left|z_n-\frac db\right|
 \le\frac{2\varepsilon}{b}.
\]
Letting $\varepsilon\downarrow0$ completes the proof.
\end{proof}

\subsection{Moments of the nonlinear and linear iterates}
We next establish second- and fourth-moment bounds for the nonlinear
iterates and their principal linear approximation.

\begin{lemma}
\label{lem:moments}
For the polynomial stepsize schedule specified above,
\begin{align}
 \PE\theta_k^2+\PE U_k^2\le C\alpha_k, \quad 
 \PE\theta_k^4+\PE U_k^4\le C\alpha_k^2.
 \label{eq:nonlinear-moments}
\end{align}
\end{lemma}
\begin{proof}
Write $A_k=1+R_k\in\{0,2\}$ and define
\[
 T_{k,A}(x)=x-\alpha_k A f'(x).
\]
Since $f'(0)=0$, $1/2\le f''\le3/2$, and
$\alpha_k A(3/2)\le3/4$, the derivative of this map satisfies
\[
 0\le T_{k,A}'(x)\le1-\frac{\alpha_k A}{2}.
\]
The mean value theorem therefore gives
\begin{equation}
 |T_{k,A}(x)|\le\left(1-\frac{\alpha_k A}{2}\right)|x|.
 \label{eq:map-contraction}
\end{equation}
For $p=2$ or $p=4$, averaging over $A_k$ yields
\begin{align*}
 \PE\left[|T_{k,A_k}(\theta_{k-1})|^p
                        \mid\filtr_{k-1}\right]
 &\le\frac{1+(1-\alpha_k)^p}{2}|\theta_{k-1}|^p\\
 &\le\left(1-\frac{\alpha_k}{2}\right)|\theta_{k-1}|^p.
\end{align*}
The update is $\theta_k=T_{k,A_k}(\theta_{k-1})-\alpha_kS_k$.
The variable $S_k$ is independent of $(A_k,\filtr_{k-1})$ and has
vanishing odd moments. Expanding the square, and then the fourth power,
thus gives
\begin{align}
 \PE\theta_k^2
 &=\PE|T_{k,A_k}(\theta_{k-1})|^2+\alpha_k^2 \leq \left(1-\frac{\alpha_k}{2}\right)\PE\theta_{k-1}^2
       +\alpha_k^2,\label{eq:second-recursion}\\
 \PE\theta_k^4
 &=\PE|T_{k,A_k}(\theta_{k-1})|^4
       +6\alpha_k^2\PE|T_{k,A_k}(\theta_{k-1})|^2
       +\alpha_k^4 \leq \left(1-\frac{\alpha_k}{2}\right)\PE\theta_{k-1}^4
       +6\alpha_k^2\PE\theta_{k-1}^2+\alpha_k^4.
 \label{eq:fourth-recursion}
\end{align}
Because $\theta_0=0$, iterating \eqref{eq:second-recursion} and
applying \Cref{lem:kernel} with $b=1/2$ and $r=1$ gives
\begin{align*}
\PE\theta_k^2 \leq \sum_{i=1}^k\alpha_i^2
       \prod_{\ell=i+1}^k\left(1-\frac{\alpha_\ell}{2}\right) \leq \sum_{i=1}^k\alpha_i^2
       \exp\left\{-\frac12\sum_{\ell=i+1}^k\alpha_\ell\right\}
 \le C\alpha_k\,.
\end{align*}
For $i\ge2$, this also gives
$\PE\theta_{i-1}^2\le C\alpha_i$; the term at $i=1$ is zero.
Substituting into \eqref{eq:fourth-recursion} and iterating once more,
\begin{align*}
\PE\theta_k^4 
 \leq C\sum_{i=1}^k(\alpha_i^3+\alpha_i^4) \exp\left\{-\frac12\sum_{\ell=i+1}^k\alpha_\ell\right\} \leq C\alpha_k^2\,.
\end{align*}
The last line uses \Cref{lem:kernel} with $r=2$ and $r=3$, and
$\alpha_k\le1/4$.

For $U_k$, independence in \eqref{eq:nonlinear-U} gives the exact
identities
\begin{align*}
 \PE U_k^2
   &=(1-\alpha_k)^2\PE U_{k-1}^2+\alpha_k^2,\\
 \PE U_k^4
   &=(1-\alpha_k)^4\PE U_{k-1}^4
        +6\alpha_k^2(1-\alpha_k)^2\PE U_{k-1}^2+\alpha_k^4.
\end{align*}
For $p=2,4$, $(1-\alpha_k)^p\le1-\alpha_k/2$.
The same two iterations prove the bounds for $U_k$.
\end{proof}

Coupling the nonlinear and linear recursions through the same noise
gives the following $L^2$ bound on the difference between $\theta_k$
and $U_k$.

\begin{lemma}\label{lem:coupling}
For the polynomial schedule,
\begin{equation}
 \norm{\theta_k-U_k}_2\le C\alpha_k.
 \label{eq:nonlinear-coupling}
\end{equation}
\end{lemma}
\begin{proof}
Subtracting \eqref{eq:nonlinear-U} from the SGD update gives
\begin{align*}
 e_k:=\theta_k-U_k
 &=(1-\alpha_k)e_{k-1}
       -\frac{\alpha_k}{2}q(\theta_{k-1})
       -\alpha_kR_k f'(\theta_{k-1}).
\end{align*}
Since $e_0=0$, iteration yields $e_k=D_k+M_k$, where we have defined
\begin{align*}
 D_k = - \frac12\sum_{i=1}^k\alpha_iP_{i,k}q(\theta_{i-1}), \quad M_k =-\sum_{i=1}^k\alpha_iP_{i,k}R_i f'(\theta_{i-1})\,.
\end{align*}
We bound these two terms separately. By \eqref{eq:q-taylor} and
\Cref{lem:moments},
\[
 \norm{q(\theta_{i-1})}_2
 \leq (1/2) (\PE\theta_{i-1}^4)^{1/2}
 \le C\alpha_i.
\]
Minkowski's inequality and \Cref{lem:kernel} consequently give
\begin{align*}
 \norm{D_k}_2 \leq \frac12\sum_{i=1}^k\alpha_iP_{i,k}
                \norm{q(\theta_{i-1})}_2 \leq C\sum_{i=1}^k\alpha_i^2P_{i,k}
 \le C\alpha_k\,.
\end{align*}
To handle $M_k$, note that $R_i f'(\theta_{i-1})$ is measurable
with respect to $\filtr_i$ and
\[
 \PE[R_i f'(\theta_{i-1})\mid\filtr_{i-1}]
 =f'(\theta_{i-1})\PE R_i=0.
\]
For $i<j$, conditioning the product on $\filtr_{j-1}$ therefore
shows that its expectation is zero. Since the weights are deterministic,
\begin{align*}
 \norm{M_k}_2^2 =\sum_{i=1}^k\alpha_i^2P_{i,k}^2
                   \PE\{f'(\theta_{i-1})^2\} \leq \frac94\sum_{i=1}^k\alpha_i^2P_{i,k}^2
                   \PE\theta_{i-1}^2 \leq C\sum_{i=1}^k\alpha_i^3P_{i,k}^2
 \le C\alpha_k^2.
\end{align*}
Here we used $|f'(x)|\le3|x|/2$, followed by
\Cref{lem:moments,lem:kernel}. Finally,
\[
\norm{e_k}_2\le\norm{D_k}_2+\norm{M_k}_2\le C\alpha_k\,.
\]
\end{proof}

The moment and coupling estimates now control the forcing errors
that enter the recursion for $H_k$.

\begin{lemma}
\label{lem:forcing}
For the polynomial schedule and every $k\ge1$,
\begin{align}
 \PE\left|q(\theta_{k-1})-\frac{U_{k-1}^2}{2}\right|
    &\le C\alpha_k^{3/2},\notag\\
 \norm{f'(\theta_{k-1})-U_{k-1}}_2
    &\le C\alpha_k.
 \label{eq:nonlinear-forcing-errors}
\end{align}
\end{lemma}
\begin{proof}
For a generic index $j\ge1$, add and subtract $\theta_j^2/2$.
Taylor's estimate \eqref{eq:q-taylor}, followed by
Cauchy--Schwarz, gives
\begin{align*}
 \PE\left|q(\theta_j)-\frac{U_j^2}{2}\right|
 &\le\PE\left|q(\theta_j)-\frac{\theta_j^2}{2}\right|
        +\frac12\PE|\theta_j^2-U_j^2|\\
 &\le\frac1{24}\PE\theta_j^4
        +\frac12\norm{\theta_j-U_j}_2\norm{\theta_j+U_j}_2\\
 &\le C\alpha_j^2+C\alpha_j\sqrt{\alpha_j}\\
 &\le C\alpha_j^{3/2}.
\end{align*}
The penultimate line uses \Cref{lem:moments,lem:coupling}.
Next, \eqref{eq:gradient-split} implies
\begin{align*}
 \norm{f'(\theta_j)-U_j}_2
 &\le\norm{\theta_j-U_j}_2
       +\frac12\norm{q(\theta_j)}_2\\
 &\le C\alpha_j+\frac14(\PE\theta_j^4)^{1/2}
 \le C\alpha_j.
\end{align*}
Taking $j=k-1$ and using the comparison of adjacent step sizes proves
the result for $k\ge2$. Both expressions vanish at $k=1$.
\end{proof}

\subsection{Bounds for the correction terms}
The next lemma bounds the multiplicative and curvature corrections
and the remaining error. It also records the conditional centering
of $J_k$.

\begin{lemma}\label{lem:corrections}
For the polynomial schedule, the processes in
\eqref{eq:nonlinear-J}--\eqref{eq:nonlinear-H} satisfy
\begin{align}
 \norm{J_k}_2+\norm{K_k}_2 \le C\alpha_k, \quad 
 \norm{H_k}_1 \le C\alpha_k^{3/2}.
 \label{eq:correction-moments}
\end{align}
In addition,
\begin{equation}
 \PE[J_k\mid S_1,\ldots,S_k]=0.
 \label{eq:J-conditional-zero}
\end{equation}
\end{lemma}
\begin{proof}
The sum representations are given in \eqref{eq:correction-sums}.
The variables $R_iU_{i-1}$ are martingale differences for
$(\filtr_i)_{i\ge0}$. Orthogonality, \Cref{lem:moments}, and then
\Cref{lem:kernel} yield
\begin{align*}
 \norm{J_k}_2^2 =\sum_{i=1}^k\alpha_i^2P_{i,k}^2\PE U_{i-1}^2 \le C\sum_{i=1}^k\alpha_i^3P_{i,k}^2
 \le C\alpha_k^2.
\end{align*}
For the curvature term, Minkowski's inequality gives
\begin{align*}
 \norm{K_k}_2 \leq \frac14\sum_{i=1}^k\alpha_iP_{i,k}\norm{U_{i-1}^2}_2 \leq C\sum_{i=1}^k\alpha_i^2P_{i,k}
 \le C\alpha_k.
\end{align*}
To bound the remainder, introduce the two forcing variables
\[
 a_i=q(\theta_{i-1})-\frac{U_{i-1}^2}{2},
 \qquad b_i=f'(\theta_{i-1})-U_{i-1}.
\]
Unrolling \eqref{eq:nonlinear-H} gives
\[
 H_k=-\frac12\sum_{i=1}^k\alpha_iP_{i,k}a_i
           -\sum_{i=1}^k\alpha_iP_{i,k}R_i b_i.
\]
The first sum is bounded in $L^1$ using \Cref{lem:forcing}:
\begin{align*}
\left\|\frac12\sum_{i=1}^k\alpha_iP_{i,k}a_i\right\|_1 \leq \frac12\sum_{i=1}^k\alpha_iP_{i,k}\PE|a_i| \leq C \sum_{i=1}^k\alpha_i^{5/2}P_{i,k}
 \le C\alpha_k^{3/2}\,.
\end{align*}
For the second sum, $b_i$ is $\filtr_{i-1}$-measurable and $R_i$
is independent of $\filtr_{i-1}$. Thus $R_i b_i$ are martingale
differences, and
\begin{align*}
 \left\|\sum_{i=1}^k\alpha_iP_{i,k}R_i b_i\right\|_2^2 =\sum_{i=1}^k\alpha_i^2P_{i,k}^2\PE b_i^2 \leq C\sum_{i=1}^k\alpha_i^4P_{i,k}^2
 \le C\alpha_k^3.
\end{align*}
Combining these estimates and $\norm{V}_1\le\norm{V}_2$ proves
the bounds in \eqref{eq:correction-moments}.
Finally, $U_{i-1}$ is a function of $S_1,\ldots,S_{i-1}$, while
$R_i$ is independent of the entire sequence $(S_j)_{j\ge1}$ and
has mean zero. Taking the conditional expectation term by term in
the representation of $J_k$ proves \eqref{eq:J-conditional-zero}.
\end{proof}

Finally, we compute the leading correction to the limiting variance
of the linearized average. This identifies the coefficient of
$n^{\gamma-1}$ in $s_n^2-1$, with $s_n^2$ defined in
\eqref{eq:pr-linear-variance}. 

\begin{lemma}
\label{lem:covariance}
For the polynomial step sizes in \eqref{eq:nonlinear-recursion} and
the weights in \eqref{eq:weights},
\begin{equation}
\label{eq:cov-expansion}
 s_n^2=\frac1n\sum_{i=1}^{n-1}Q_{i,n}^2
      =1+2n^{\gamma-1}+o(n^{\gamma-1}).
\end{equation}
\end{lemma}
\begin{proof}
The weights $Q_i$ in the lower-bound construction of
\citet[Appendix~E]{sgd} coincide with $Q_{i,n}$ for $G=1$,
$c_0=1/4$, and $k_0=1$. Appendix~E of \citet{sgd} gives 
\begin{equation}
\label{eq:pr-cited-covariance-reduction}
 s_n^2-1 = \frac{L_{n-1}}n+O(n^{2\gamma-2}+n^{-1}),
 \qquad \text{where }
 L_k=\sum_{i=1}^k\prod_{\ell=i}^k(1-\alpha_\ell)^2.
\end{equation}
These estimates depend only on the deterministic weights. It remains
to compute the leading coefficient of $L_{n-1}/n$. Since $L_0=0$,
the definition of $L_n$ gives
\[
 L_n=(1-\alpha_n)^2(1+L_{n-1}).
\]
For $n \geq 2$, putting $u_n=\alpha_n L_n$ implies that
\begin{align*}
 u_n = (1-\alpha_n)^2\frac{\alpha_n}{\alpha_{n-1}}u_{n-1} +\alpha_n(1-\alpha_n)^2 = (1-2\alpha_n+o(\alpha_n))u_{n-1} +\alpha_n(1+o(1))\,.
\end{align*}
Here we used the fact that 
\[
\alpha_n/\alpha_{n-1}=1+O(n^{-1})=1+o(\alpha_n)\,.
\]
By \Cref{lem:tracking}, $u_n \to 1/2$. Consequently,
\[
 \frac{L_{n-1}}n
 =\frac{1+o(1)}{2n\alpha_{n-1}}
 =2n^{\gamma-1}+o(n^{\gamma-1}),
\]
and substitution into
\eqref{eq:pr-cited-covariance-reduction} implies \eqref{eq:cov-expansion}.
\end{proof}

\end{document}